\documentclass[letterpaper, 10 pt, conference]{ieeeconf}  % Comment this line out if you need a4paper

\IEEEoverridecommandlockouts                              % This command is only needed if 
\usepackage{times}
\usepackage{microtype}
\usepackage{amsmath,amsfonts}
\usepackage{mathtools}
\usepackage{float}
\usepackage{textcomp}
\usepackage{multicol}
\usepackage{multirow}
\usepackage{balance}
\usepackage[short,c2]{optidef}

\newcommand{\secref}[1]{\S\ref{#1}}

\newtheorem{remark}{Remark}
\usepackage{makeidx}  % allows for index generation
\makeindex
\usepackage{xcolor}
\usepackage{xspace}
\usepackage{colortbl}
\usepackage{placeins}
\usepackage{booktabs}
\usepackage{siunitx}
\usepackage{algorithm}
\usepackage{algpseudocode}
\usepackage{bm} 
\usepackage{cuted} 
\usepackage{capt-of}
\usepackage{subcaption} 
\usepackage{xcolor}
\usepackage[normalem]{ulem}
\usepackage{booktabs}
\usepackage{tabularx}
\usepackage{array}
\definecolor{mycitecolor}{RGB}{71, 191, 38}
\definecolor{mylinkcolor}{RGB}{40, 115, 201}
\makeatletter
\let\NAT@parse\undefined
\makeatother
\usepackage[bookmarks=true, colorlinks, citecolor=mycitecolor,linkcolor=mylinkcolor,urlcolor=mycitecolor]{hyperref}

\newcommand{\method}{ALTER\xspace}

\newcommand{\pibase}{\pi_{\mathrm{base}}}
\newcommand{\Dbase}{D_{\mathrm{base}}}

\newcommand{\para}[1]{\smallskip\noindent\textbf{#1}}

\title{\Large 
Residual Denoising Enables Sample-Efficient Multi-Agent Coordination on Demand
}

\author{Dayi Dong, Maulik Bhatt, Aayushi Shrivastava, Lasse Peters, and Negar Mehr% <-this % stops a space
\thanks{All authors are with the Department of Mechanical Engineering, University of California Berkeley, Berkeley, CA 94709, USA {\tt \small \{dayi.dong, maulikbhatt, aayushis, lasse.peters, negar\}@berkeley.edu}}%
\thanks{This work was supported by the National Science Foundation under Grants ECCS-2438314 (CAREER Award) and CNS-2529645. The authors also thank the Office of Naval Research Young Investigator Program (ONR YIP) for support. This research was made possible by GPU resources provided via the NVIDIA Academic Grant.
}
}
\begin{document}

\IEEEaftertitletext{\vspace{-3\baselineskip}}
\maketitle
% \begingroup
% \renewcommand\thefootnote{\fnsymbol{footnote}}
% \footnotetext[1]{Indicates equal contribution.}
% \endgroup

\thispagestyle{empty}
\pagestyle{empty}

%%%%%%%%%%%%%%%%%%%%%%%%%%%%%%%%%%%%%%%%%%%%%%%%%%%%%%%%%%%%%%%%%%%%%%%%%%%%%%%%

% \begin{figure*}[t]
%   \centering
%   \includegraphics[width=\textwidth]{figs/fig1_v3.pdf}
%   \caption{\textbf{Multi-agent adaptation with \method.}
%   \method adapts a frozen single-arm visual diffusion policy to coordinated multi-agent tasks without access to the original expert demonstrations. Blue denotes the frozen source policy and its policy-distilled rollouts, while green denotes the multi-agent demonstrations and trainable coordination head. The adapted policy coordinates both robots while retaining the original single-arm skills.}
%   \label{fig:fig1}
% \end{figure*}
\begin{strip}
% \vspace{-2em}
  \centering
  \includegraphics[width=\textwidth]{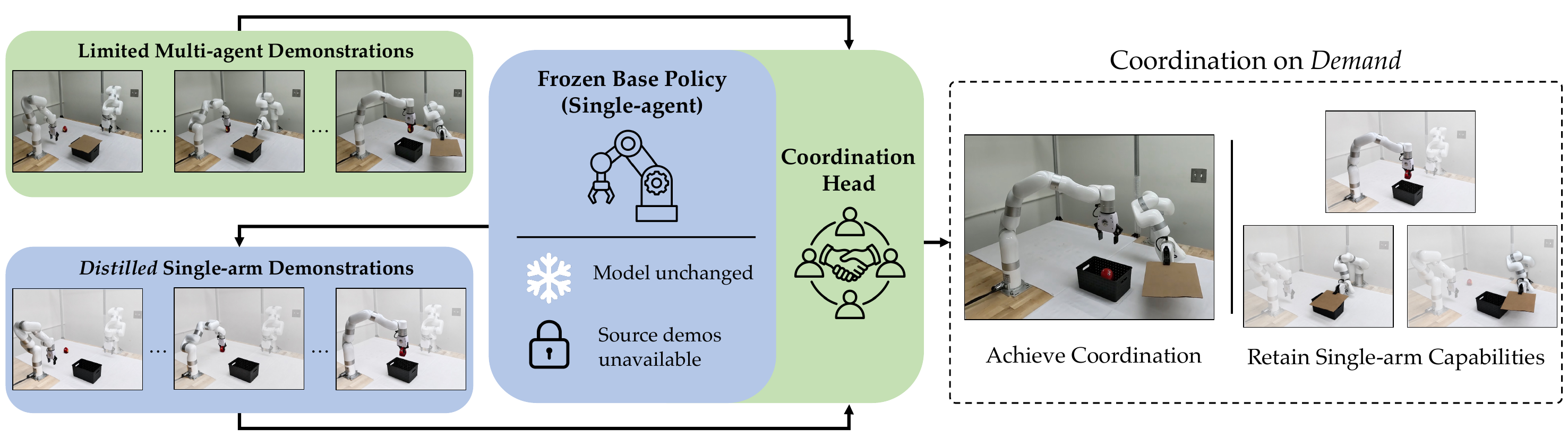}
  \captionof{figure}{\textbf{Multi-agent adaptation with \method.}
  \method adapts a frozen single-arm visual diffusion policy to coordinated
  multi-agent tasks without access to the original expert demonstrations.
  Blue denotes the frozen source policy and its policy-distilled rollouts,
  while green denotes the multi-agent demonstrations and trainable
  coordination head. Each robot runs the adapted policy using only its own
  visual observations, enabling decentralized coordination without explicit
  inter-agent communication while retaining the original single-arm skills.}
  \label{fig:fig1}
\end{strip}

\begin{abstract}
  Pretrained robot policies offer strong manipulation skills but are typically limited to single-agent settings, where a robot acts in isolation.
  In this work, we study how to adapt pretrained single-agent diffusion policies to multi-agent settings using minimal collaborative data, co-optimizing for two key objectives: high coordination performance and single-agent skill retention.
  To this end, we introduce \method, an adaptation method for \emph{coordination on demand}: the adapted policy coordinates with other robots when deployed in a team while remaining capable of acting independently when operating alone.
  Execution is decentralized: each robot acts only on its own visual observations, without explicit inter-agent communication.
  Our method trains a \emph{coordination head} that predicts a \emph{residual} denoiser to transform single-agent behavior into coordinated multi-agent behavior when necessary while also preserving single-agent capabilities.
  To preserve single-agent capabilities, we augment a small number of collaborative demonstrations with self-distilled data generated by the base policy during training of the residual denoiser.
  In simulation, \method achieves higher coordination success over our baselines while retaining much higher source-skill retention. In our hardware experiments, we find similar trends where \method better co-optimizes for coordination success and single-agent skill retention than the baselines. 
  % Hardware experiments further show that \method achieves improved coordination success while better preserving source skills.
  
  % We evaluate our approach on a suite of collaborative manipulation tasks in both simulation and hardware experiments and find that \method leads to improved coordination success and single-agent skill retention compared to training-from-scratch and full-policy-finetuning baselines.
\end{abstract}

% Reserve the first column for text; place the front figure atop the second.
% \suppressfloats[t]
% \begin{figure}[t]
%   \centering
%   \includegraphics[width=\columnwidth]{figs/fig1_v2.pdf}
%     \caption{\textbf{Multi-agent adaptation with \method.}
%     \method adapts a frozen single-arm visual diffusion policy to coordinated multi-agent tasks without access to the original expert demonstrations. Blue denotes the frozen source policy and its policy-distilled rollouts, while green denotes the multi-agent demonstrations and trainable coordination head. The adapted policy coordinates both robots while retaining the original single-arm skills.}
%   \label{fig:fig1}
% \end{figure}

%%%%%%%%%%%%%%%%%%%%%%%%%%%%%%%%%%%%%%%%%%%%%%%%%%%%%%%%%%%%%%%%%%%%%%%%%%%%%%%%

\section{Introduction} \label{sec:intro}
% THE WHY: Why should I care?
Imagine you have just enjoyed a dinner party with friends and now face the less delightful task of clearing the table.
Luckily, you have two state-of-the-art robots at home to help.\footnote{You could, of course, have asked your friends to help clean up. Why you did not is left to the reader's imagination.}
To be useful in this setting (beyond dancing~\cite{sun2026robotdancing} or performing roundhouse kicks~\cite{han2025kungfubot2} in a corner), your robots must \emph{manipulate} their environment: \emph{change} the world around them by picking up objects, opening cabinets, putting objects away, and wiping the table.

% THE OBSTACLE: Why is this hard?
In principle, generative policies based on flow matching or diffusion models enable robots to learn such skills from demonstrations~\cite{chi2025diffusion}.
In fact, numerous recent works have shown great success in training individual agents to perform complex manipulation skills using this approach~\cite{intelligence2025pi_,octo2024octo,liu2025rdt}, but these results remain largely confined to the single-agent domain.
Enabling \emph{multiple} robots to collaborate in a \emph{decentralized} fashion could offer a cost-effective way to scale their productivity.
\footnote{Much as sharing the task with your friends would have helped, had you asked.}
However, extending imitation learning (IL) to this multi-agent setting is challenging because collecting multi-agent data at scale is costly.
%On the other hand, numerous recent works on diffusion- or flow-based imitation learning have shown great success in single-agent settings~\cite{intelligence2025pi_,octo2024octo,liu2025rdt}. 
% These works have publicly available checkpoints that we could use off-the-shelf.

% THE WHAT: What is the problem that we are solving?
In view of the multi-agent data bottleneck, and the fact that there exist strong off-the-shelf policies for single-agent settings, rather than learning coordination behavior from scratch, in this work we treat multi-agent learning as an adaptation problem, starting from a pretrained image-conditioned single-agent diffusion policy, which we call the \emph{base policy} from now on.
%The source task is the single-agent setting in which the base policy was trained, and the target task is the multi-agent coordination setting we seek success in.
Starting from such a base policy, we ask
\begin{quote}
    \emph{How can we adapt a single-agent diffusion policy to coordinate with others using limited multi-agent data while retaining its single-agent capabilities?}
\end{quote}
To make our contribution applicable to off-the-shelf policies, we study this question without assuming access to the base policy's training data.

% THE HOW: On a very high level, how do we achieve this?
Our main contribution, \emph{\method} 
(\textbf{A}daptation from \textbf{L}imited demonstrations for \textbf{T}eam coordination with \textbf{E}xisting-skill \textbf{R}etention),
% \footnote{\textbf{A}daptation from \textbf{l}imited demonstrations for \textbf{t}eam coordination with \textbf{e}xisting-skill \textbf{r}etention} 
is a multi-agent adaptation paradigm that tackles precisely this challenge.
At a high level, our approach comprises three stages (cf. Fig.~\ref{fig:fig1}):
(\romannumeral1)~we roll out the base policy to generate demonstrations of single-agent skills (in Fig.~\ref{fig:fig1}: picking and placing objects);
(\romannumeral 2)~the user provides a limited number of multi-agent demonstrations of coordinated behavior (in Fig.~\ref{fig:fig1}: opening the box with one robot for another robot to place an object inside);
(\romannumeral 3)~to achieve multi-agent coordination while retaining single-agent skill during adaptation, we combine the self-distillation data from step~1 with the multi-agent data from step~2 to train a \emph{residual coordination head} that bridges the gap between the frozen single-agent policy and the target coordinated behavior.
At deployment, each robot runs its own copy of the adapted policy using only its local visual observations, enabling decentralized coordination without explicit inter-agent communication.

% THE PROPERTIES:
We compare our method to several imitation learning baselines across a suite of collaborative manipulation tasks both in simulation and in hardware.
Our results show that \method adapts base policies more effectively than the baseline approaches: it achieves higher coordination success while ensuring better retention of single-agent skills (i.e., the ability to act independently even after adaptation).
% (the ability to act on its own, even after adaptation).
%By contrast, we find that naive fine-tuning or adaptation on multi-agent data destroys single-agent capabilities, leaving agents unable to work independently.
%Beyond extensive simulation experiments, we demonstrate our approach on collaborative manipulation tasks with two xArm7 robots.

% \paragraph{Contributions.}
% The principal contributions of this work are:
% \begin{itemize}
%     \item We introduce a training paradigm for augmenting pretrained single-agent diffusion policies with decentralized multi-agent coordination capabilities from limited collaborative data while retaining source-task behavior without access to the original expert demonstrations.

%     \item We develop a residual diffusion adaptation mechanism that combines a frozen source policy with policy self-distillation to preserve single-agent behavior while achieving coordination with other agents. 
% \end{itemize}

\section{Related Work}\label{sec:related}

    We position our work relative to diffusion visuomotor policies, multi-agent imitation learning, and pretrained-policy adaptation.

    \para{Diffusion Models} provide a flexible policy class for modeling multimodal action distributions~\cite{ho2020denoising,janner2022planning,chi2025diffusion}.
    Diffusion Policy~\cite{chi2025diffusion} generates action chunks from visual observations via conditional denoising.
    DP3~\cite{ze20243d} extends this formulation by conditioning on sparse 3D point clouds.
    These methods typically train a policy directly for the target task and do not directly support adaptation to new settings without retraining.
    % We instead keep a pretrained single-arm diffusion policy frozen and train only a residual head to extend it to the multi-agent coordination target task while preserving the single-arm source task behavior.

    \para{Multi-Agent Imitation Learning} aims to coordinate behavior directly from collaborative demonstrations. MIMIC-D~\cite{dong2025mimicd} jointly trains decentralized diffusion policies with a shared loss.
    CHORUS~\cite{doshi2026chorus} adapts a shared vision-language-action policy for decentralized multi-embodiment teams. Such methods are typically data-hungry and train directly for collaboration rather than preserving pretrained single-agent capabilities. 
    % fail to retain capabilities of single-agent settings.
    % Compared to these two methods, which target coordinated behavior solely from coordination data and are data hungry and task specific, our formulation utilizes a pretrained base policy for data efficiency and requires the adapted policy to retain source task competence. 
    Other diffusion approaches add explicit teammate models or consensus latent representations to support coordination~\cite{zhu2024madiff,he2025latent}, which require additional teammate-modeling or latent-consensus machinery. 
    % but our residual head executes using each agent's ego observation without these components.
    CoDi~\cite{peters2026coordinated}, on the other hand, coordinates pretrained single-agent diffusion policies without collaborative demonstrations, but instead applies a user-specified multi-agent cost at sampling time for guidance. However, it can be difficult to specify such a cost function for complex multi-agent coordination.
    % By contrast, our approach learns coordination directly from a limited set of multi-agent demonstrations.
    % Furthermore, our approach explicitly aims to preserve single-arm behavior.

    \para{Adaptation without Forgetting.}
    Our problem can be viewed as a variant of continual learning~\cite{auddy2023continual, liu2025skill} but with distinct constraints: we do not have the source data and we have highly skewed data budgets in different domains (single- vs multi-agent data).
    A second approach to adaptation without forgetting is runtime policy selection. Mixture-of-experts models and language-guided skill planners select an expert or temporally extended skill at runtime~\cite{ahn2022saycan, huang2024mentor}. In our deployment setting, however, no external signal indicates whether the robot should execute a source task or collaborative policy.
    % Our adaptation setting requires that we add coordination to a frozen single-arm policy while preserving its source task behavior without access to its source expert dataset.

    \para{Positioning of Our Approach.}
    Residual policy learning~\cite{johannink2019residual} provides a mechanism for adapting policies by combining a fixed nominal controller with a learned corrective policy. We apply the principle of residual learning at the \emph{denoiser} level.
    Our architecture is also related to parameter-efficient adaptation methods like side-tuning~\cite{zhang2020side}, which adapts a fixed pretrained network through an additive trainable side network, and ControlNet~\cite{zhang2023adding}, which augments a frozen diffusion model with zero-initialized trainable branches for additional conditioning.
    Using this idea, we keep a pretrained single-arm diffusion policy frozen and train only a residual head to extend it to the multi-agent coordination target task while preserving the single-arm source task behavior.

\section{Preliminaries: Diffusion Models}\label{subsec:diffusion-prelim}

Our method builds on conditional diffusion models, which we briefly review in the continuous-noise-level parameterization of Karras et al.~\cite{karras2022elucidating}, commonly referred to as EDM. We refer the reader to~\cite{ho2020denoising, song2021scorebased, karras2022elucidating} for a complete overview.

Diffusion models aim to generate samples from a distribution $p(x\mid c)$ over data $x$ conditioned on context $c$.
Importantly, the underlying data distribution is unknown.
Instead, the generative model must be learned from only a dataset of examples $x \sim p(\cdot \mid c)$.
Diffusion models achieve this by considering a family of noise-corrupted distributions $p_\sigma(x_\sigma \mid c) := \int p(x \mid c)\,\mathcal{N}(x_\sigma; x, \sigma^2 I)\,\mathrm{d}x$ obtained by perturbing data $x \sim p(\cdot \mid c)$ with Gaussian noise of standard deviation $\sigma \ge 0$. %, that is, $x_\sigma = x + \sigma \epsilon$ with $\epsilon \sim \mathcal{N}(0, I)$.
At $\sigma = 0$, this family coincides with the data distribution; at a sufficiently large noise level $\sigma_{\max} \gg 0$, it is indistinguishable from the Gaussian $\mathcal{N}(0, \sigma_{\max}^2 I)$, which is trivial to sample from.

The key idea of diffusion models is to generate data samples from Gaussian noise by reversing the noise-corruption process.
Starting from a sample $x_{\sigma_{\max}} \sim \mathcal{N}(0, \sigma_{\max}^2 I)$, we denoise it into a clean sample by integrating the following probability flow ordinary differential equation (ODE)~\cite{song2021scorebased}
% Sampling proceeds by transporting Gaussian noise to data. Starting from $x_{\sigma_{\max}} \sim \mathcal{N}(0, \sigma_{\max}^2 I)$, the model integrates the probability-flow ordinary differential equation (ODE)~\cite{song2021scorebased,karras2022elucidating}
\begin{equation}
    \frac{\mathrm{d} x_\sigma}{\mathrm{d} \sigma}
    =
    -\sigma\, s(x_\sigma, \sigma, c)
    \label{eq:pf-ode}
\end{equation}
from $\sigma = \sigma_{\max}$ down to $\sigma = 0$, where $s(x_\sigma, \sigma, c) := \nabla_{x_\sigma} \log p_\sigma(x_\sigma \mid c)$ is the score function of the noised distribution. Intuitively, the score function points towards high-probability regions of the underlying data distribution $p(x\mid c)$. 
% From now on, we use $s(x_\sigma,\sigma,c):= \nabla_{x_\sigma} \log p_\sigma(x_\sigma \mid c)$ to denote the score function. Integrating the ODE yields a sample from $p(x \mid c)$.

Since the score function $s_\sigma$ in~\eqref{eq:pf-ode} is unknown, diffusion models learn it from data.
In the EDM formulation, this is achieved by training a denoiser $D_\theta(x_\sigma, \sigma, c)$ with parameters $\theta$, a network that predicts the clean $x$ from its noised version, by denoising regression using the following loss function
\begin{equation}
    \mathcal{L}_{\mathrm{den}}(\theta; \mathcal{D})
    =
    \mathbb{E}_{\mathcal{D},\, \sigma,\, \epsilon}
    \left[
        w(\sigma)
        \left\|
            D_\theta(x + \sigma \epsilon, \sigma, c) - x
        \right\|_2^2
    \right],
    \label{eq:denoising-loss}
\end{equation}
where the expectation is over $(x, c) \sim \mathcal{D}$, $\sigma \sim p(\sigma)$, and $\epsilon \sim \mathcal{N}(0, I)$, and the noise-level distribution $p(\sigma)$ and the weighting $w(\sigma) > 0$ are design choices of the training recipe~\cite{karras2022elucidating}. By Tweedie's formula~\cite{efron2011tweedie,karras2022elucidating}, the minimizer $D^\star$ of~\eqref{eq:denoising-loss} is the posterior mean $\mathbb{E}[x \mid x_\sigma, c]$ and satisfies
\begin{equation}
    D^\star(x_\sigma, \sigma, c)
    =
    x_\sigma + \sigma^2 s(x_\sigma, \sigma, c).
    \label{eq:score-identity}
\end{equation}
Therefore, any learned denoiser $D_\theta$ induces a score estimate $s_{D_\theta}(x_\sigma, \sigma, c) := \left( D_\theta(x_\sigma, \sigma, c) - x_\sigma \right) / \sigma^2$ that we can use to estimate data samples via~\eqref{eq:pf-ode}. In this work, we solve the ODE \eqref{eq:pf-ode} numerically using Algorithm~1 in~\cite {karras2022elucidating}.

% \begin{remark}
% Note that denoisers can be combined by addition. If we add a correction term $\Delta(x_\sigma, \sigma, c)$ to a denoiser $D$, the identity above gives the sum an exact meaning in score space: $s_{D + \Delta} = s_D + \Delta / \sigma^2$, so the correction shifts the estimated score field by a $\sigma^{-2}$-rescaled amount at every noise level. The combined denoiser $D + \Delta$ therefore plugs into the sampler of~\eqref{eq:pf-ode} with no change to the noise schedule or the solver. This property is the basis of our method.
% \end{remark}

\section{Problem Formulation}\label{sec:prob}

In this section, we formalize the problem of adapting a pretrained single-arm policy for multi-arm coordination. 
% We model collaborative manipulation as a decentralized partially observable Markov decision process (Dec-POMDP) and state how we represent policies as diffusion models (\secref{subsec:dec-pomdp}), and we then specify the adaptation setting together with our adaptation objective (\secref{subsec:adaptation-setting}).

    % \subsection{Decentralized Multi-Agent Imitation Learning}\label{subsec:dec-pomdp}

\para{Task Model.}
We consider a collaborative task between $N$ homogeneous agents over a finite horizon of $T$ steps.
Throughout this work, we use the superscript $i$ and the subscript $t$ to denote a quantity associated with agent $i$ at time $t$.
The agents operate in a shared workspace and perceive it only through their own sensors. 
Hence, we model the collaborative task as a Dec-POMDP, $\mathcal{M} = \langle \mathcal{S}, \mathcal{A}, \mathcal{O}, P, Z, \rho_0, N \rangle$~\cite{oliehoek2016concise}. Here, $\mathcal{S}$ is the global state space of the workspace, which includes all agents and objects; $\mathcal{A} = \mathcal{A}^1 \times \cdots \times \mathcal{A}^N$ is the joint action space, where $\mathcal{A}^i = \mathbb{R}^{d_a}$ is the action space of agent $i$ and $d_a$ is the action dimension of each agent; $\mathcal{O} = \mathcal{O}^1 \times \cdots \times \mathcal{O}^N$ is the joint observation space, where $\mathcal{O}^i \subseteq \mathbb{R}^{d_o}$ is the continuous local observation space of agent $i$ and $d_o$ is the per-agent observation dimension; $P$ is a Markov transition kernel where for every global state $s \in \mathcal{S}$ and joint action $a_t^{1:N} := (a^1_t, \dots, a^N_t) \in \mathcal{A}$, $P(\cdot \mid s, a^{1:N})$ is a probability distribution over the next global state in $\mathcal{S}$; $Z = (Z^1, \dots, Z^N)$ collects the per-agent observation kernels $Z^i(\cdot \mid s)$ which are probability distributions over the local observation space $\mathcal{O}^i$ given global state $s$; and $\rho_0$ is the initial-state distribution. 
% In our POMDP tuple, we omit a reward function because we want the agents to learn directly from demonstrations. %; we evaluate policies through task success, as we make precise in \secref{subsec:adaptation-setting}.

\para{Decentralized Execution.} Centralized planning and explicit inter-agent communication can be unreliable or unavailable at deployment. %, especially when robots must operate independently or alongside partners that they cannot directly communicate with.
Therefore, we require each agent to make decisions based only on its own local observation while coordinating with other agents sharing the same workspace.
Assuming that all agents are driven by the same policy, i.e., $\pi^i = \pi$, we model the joint policy under this decentralized execution assumption as
% \footnote{Note that sharing policy parameters does not weaken decentralization because each agent still only receives their own observations.}
\begin{equation}
\pi^{1:N}(a^{1:N} \mid o^{1:N})
:=
\prod_{i=1}^{N} \pi^i(a^i \mid o^i)
=
\prod_{i=1}^{N} \pi(a^i \mid o^i).
\label{eq:factorized-joint-policy}
\end{equation}
Note that sharing policy parameters does not weaken decentralization because each agent receives only its own observations. In summary, starting from initial states $s_0$ drawn from  $\rho_0$, at every time step, each agent receives an observation and selects an action, and the joint action of all agents drives the workspace to its next state according to the following closed-loop stochastic dynamics
\begin{equation}
o^i_t \sim Z^i(\cdot \mid s_t), \;
a^i_t \sim \pi(\cdot \mid o^i_t),
\;
s_{t+1} \sim P(\cdot \mid s_t, a^{1:N}_t).
\label{eq:dec-pomdp-dynamics}
\end{equation} %The process in~\eqref{eq:dec-pomdp-dynamics} encodes the decentralized execution constraint that we impose throughout this work. 

\para{Diffusion Policies with Action Chunking.} We represent the policy $\pi$ as a conditional diffusion model as introduced in Sec.~\ref{subsec:diffusion-prelim}, where the distribution now is over an agent's \emph{actions} (the \emph{data} $x$ in \secref{subsec:diffusion-prelim}) conditioned on local RGB images of the scene (the \emph{context} $c$ in \secref{subsec:diffusion-prelim}).
As is common practice~\cite{chi2025diffusion}, we instantiate diffusion policies with action chunking~\cite{chi2025diffusion,zhao2023learning}: at time $t$, agent $i$ predicts an $H$-step action chunk $a^i_{t:t+H}:= (a^i_t, \dots, a^i_{t+H-1}) \in \mathbb{R}^{H \times d_a}$ from its observation $o^i_t$, where $H$ is the prediction horizon, executes the first $H_a \le H$ actions, and then replans from its next observation. With a slight abuse of notation, we therefore let $\pi(\cdot \mid o^i_t)$ denote a distribution over action chunks $a^i_{t:t+H}$, and the dynamics~\eqref{eq:dec-pomdp-dynamics} execute chunks in this receding-horizon fashion.

\para{Collaborative Demonstrations.} We denote by $\mathcal{D}_C$ a set of $M_C$ expert episodes in which all $N$ agents perform a collaborative task together,
\begin{equation}
\mathcal{D}_C = \left\{ \tau^C_m \right\}_{m=1}^{M_C},
\qquad
\tau^C_m = \left\{ \left( o^{1:N}_t, a^{1:N}_t \right) \right\}_{t=0}^{T-1},
\label{eq:collab-data}
\end{equation}
where each episode $\tau^C_m$ records the joint observations $o^{1:N}_t$ and joint actions $a^{1:N}_t$ at every time step.

\para{The Adaptation Problem: Coordination on Demand.}
The goal of decentralized multi-agent imitation learning is to learn a homogeneous policy $\pi$ such that when each agent employs a copy of $\pi$ in a decentralized fashion, the joint behavior under~\eqref{eq:dec-pomdp-dynamics} reproduces the coordinated behavior in $\mathcal{D}_C$.
Rather than learning coordinated behavior from scratch based on only $\mathcal{D}_C$, in this work, we assume access to a \emph{pretrained policy} $\pibase$ (represented as a diffusion model with a corresponding denoiser $\Dbase$) that we will use as a starting point for adaptation.
Going forward, we shall refer to the single-agent tasks that this base policy was trained to perform as \emph{source tasks} and call the target multi-agent tasks in $\mathcal{D}_C$ the \emph{collaborative tasks}. 
% a partner is present and no external signal that selects a single-arm or multi-arm mode.
% Therefore, agent $i$ must decide from its local observation $o_t^i$ alone, as in~\eqref{eq:dec-pomdp-dynamics}. This rules out an explicit runtime switch between separate source and collaborative policies. Such a switch would require inferring partner presence from the same partial observation, which is ambiguous in our setting and difficult to calibrate reliably (\secref{sec:related}).            

We consider deployment conditions in which a robot may operate alone in some episodes and alongside partners in others and assume that each agent receives no label indicating whether collaboration is required (i.e., whether the current episode features one or multiple agents).
Therefore, we seek to train an adapted policy that can seamlessly transition between single-agent skills and multi-agent coordination on demand.
% A single adapted policy must autonomously switch between source-task behavior when operating alone and coordinated behavior when collaboration is required.
Hence, we must adapt $\pibase$ to the collaborative (multi-agent) setting while preserving its source-task (single-agent) capabilities.

% To do so, we keep $\Dbase$ frozen and learn additional trainable components for coordination. During adaptation, we use the collaborative demonstrations $\mathcal{D}_C$ from~\eqref{eq:collab-data} and execute $\pibase$ in the source-task environment to collect source-policy rollouts.

\para{Challenges of Multi-Agent Adaptation.}
The adaptation of a pretrained single-agent policy to a collaborative multi-agent setting is complicated by two main challenges.
%A natural way to learn coordination is to perform multi-agent imitation learning on collaborative demonstrations $\mathcal{D}_C$.
First, a key challenge is that the collaborative demonstrations $\mathcal{D}_C$ are limited in number since collecting multi-agent demonstrations at scale is difficult in practice due to the need to operate multiple robots simultaneously.
Therefore, we aim to learn coordinated behavior from a small number of multi-agent demonstrations, i.e., in the low-data regime.

Second, while we have access to a pretrained policy $\pibase$, we may not have access to the original expert demonstrations that were used to train it.
This assumption reflects a common deployment scenario in which robot policies are distributed as checkpoints without their training demonstrations~\cite{intelligence2025pi_,octo2024octo,liu2025rdt}.
Without the original expert data, we cannot rehearse the source dataset alongside new collaborative data, which is the standard defense against catastrophic forgetting~\cite{robins1995catastrophic}. Any supervision about source-task behavior must therefore come from $\pibase$ itself.

\section{\method{}} \label{subsec:codiff}

We now present our main contribution, \method{} (\textbf{A}daptation from \textbf{L}imited demonstrations for \textbf{T}eam coordination with \textbf{E}xisting-skill \textbf{R}etention), to tackle the adaptation problem outlined in~Sec.~\ref{sec:prob}.
Fig.~\ref{fig:fig1} illustrates the high-level approach of our method, which rests on three key decisions:
\begin{enumerate}
    \item First, we keep the pretrained denoiser $\Dbase$ (parameterizing $\pibase$) frozen and add a small residual coordination adapter on top of it so that we can maximize the use of skills already learned by $\Dbase$ while learning coordination (\secref{sec:residual_adapter}).
    \item Second, we replace the missing source dataset with \emph{policy-distilled replay}, a dataset of rollouts of the frozen policy in its source environment (\secref{sec:mixed_training}). 
    \item Third, we train the adapter on the collaborative demonstrations and the replay data jointly, so that a single set of parameters serves both source and target tasks at deployment (\secref{sec:mixed_training}).
\end{enumerate}
Below, we provide more details on each one of these components.
% TODO(review): fig:fig1 caption is still a placeholder; the sentence above assumes it shows replay collection, mixed-domain training, and online execution.
% NOTE: The Policy Representation and Architecture subsection moved, condensed, into the Implementation Details subsection at the end of this section (Lasse's comment).

\begin{figure}[t]
    \centering
    \includegraphics[width=0.98\linewidth]{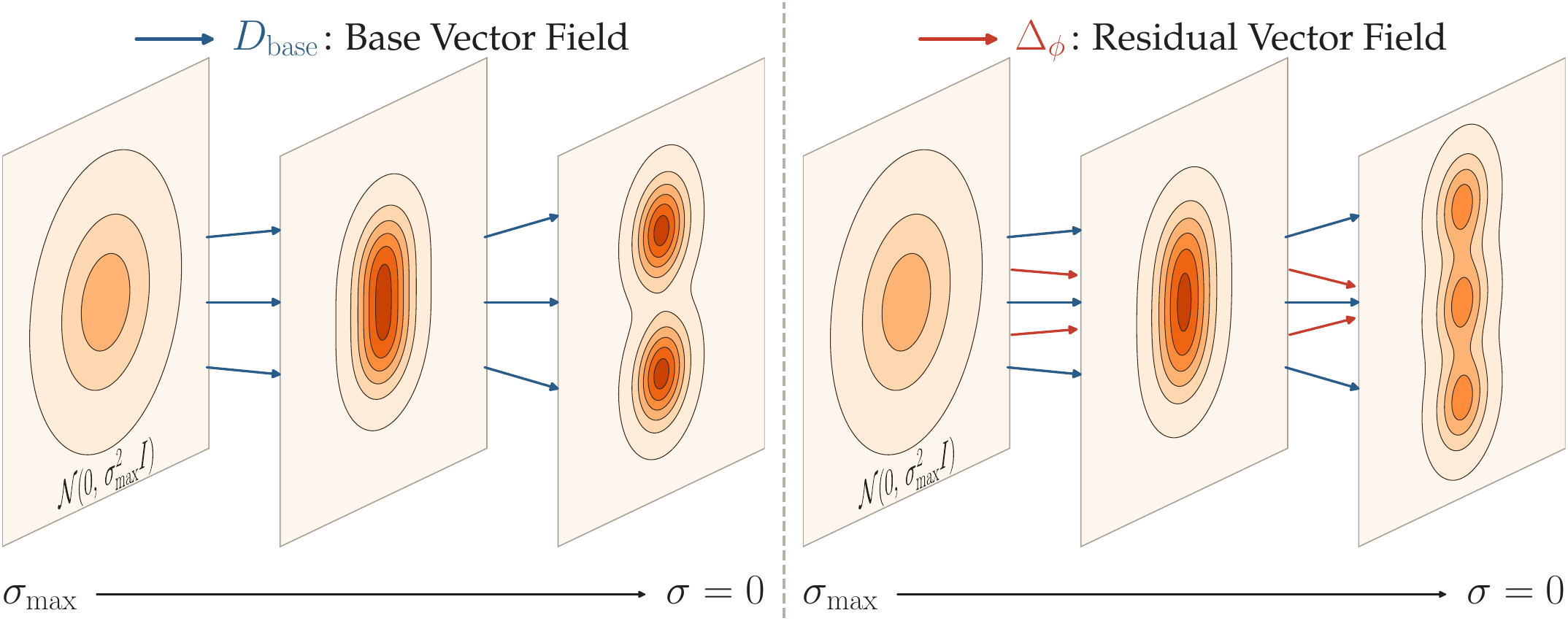}
    \caption{{\textbf{The residual as a correction to the base vector field.} \emph{Left:} the frozen base field of $\Dbase$ (blue) alone yields the action distribution of $\pibase$. \emph{Right:} adding the residual field $\Delta_\phi$ (red) at every noise level bends the same trajectories toward the coordinated action distribution.}}
    \vspace{-0.5cm}\label{fig:residual-field}
\end{figure}

\subsection{Residual Coordination Adapter} \label{sec:residual_adapter}

The pretrained denoiser already provides competent per-arm behavior; what it lacks is the partner-dependent adjustment that turns individually competent robots into a coordinated team. We therefore leave $\Dbase$ untouched and train a \emph{residual coordination adapter} $\Delta_\phi$, a second small denoising network with trainable parameters $\phi$, whose output corrects the prediction of the frozen base. The adapter receives the same noised chunk, noise level, and observation as the base denoiser, and additionally the base prediction itself since the quantity it must produce is precisely a correction to the base prediction.
{Crucially, as the residual adapter receives only agent $i$'s local observation $o^i$, the same input as $\Dbase$, decentralized execution as in~\eqref{eq:dec-pomdp-dynamics} is preserved by construction.} Thus, the adapted denoiser sums the two outputs,
\begin{equation}
    \begin{split}
    D_{\phi}(x_{\sigma}, \sigma, o)
    =
    & \Dbase(x_{\sigma}, \sigma, o)
    \\
    & \quad +
    \Delta_{\phi}
    \left(
        x_{\sigma}, \sigma, o,
        \Dbase(x_{\sigma}, \sigma, o)
    \right).
    \end{split}
    \label{eq:adapted_denoiser}
\end{equation}
The adapted policy $\pi_\phi$ is a diffusion policy that samples action chunks through ODE~\eqref{eq:pf-ode} with $D_\phi$ as its denoiser. To simplify the notation, in the remainder of the paper, we abbreviate the residual by $\Delta_\phi(x_\sigma, \sigma, o)$, leaving its dependence on the base prediction implicit.

% \begin{proposition}[Additive Property of Residual Denoisers]
% \label{prop:residual-score}

% Let $D_\phi = \Dbase + \Delta_\phi$ as in~\eqref{eq:adapted_denoiser}, where
% $\Delta_\phi$ may depend on $(x_\sigma, \sigma, o)$ arbitrarily, including through
% $\Dbase(x_\sigma, \sigma, o)$. Then, for every $(x_\sigma, \sigma, o)$:
% \begin{enumerate}
%     \item the induced scores satisfy
% \begin{equation}
%     s_{D_\phi}(x_\sigma, \sigma, o)
%     =
%     s_{\Dbase}(x_\sigma, \sigma, o) + \sigma^{-2}\, \Delta_\phi(x_\sigma, \sigma, o),
%     \label{eq:residual-score}
% \end{equation}
% so the probability-flow ODE~\eqref{eq:pf-ode} under $s_{D_\phi}$ is that of $\Dbase$
% with the additional drift $-\Delta_\phi(x_\sigma, \sigma, o)/\sigma$;
% \item if $\Delta_\phi(\cdot, \cdot, o) \equiv 0$ for a given observation $o$,
% then $\pi_\phi(\cdot \mid o) = \pibase(\cdot \mid o)$. % for any sampler that accesses the
% % denoiser only through its evaluations, in particular Algorithm~1
% % of~\cite{karras2022elucidating}.
% \end{enumerate}
% \end{proposition}
% \begin{proof}
% Property (i) is the linearity of $D \mapsto (D - x_\sigma)/\sigma^2$, followed by substitution
% into~\eqref{eq:pf-ode}. For (ii), $D_\phi(\cdot, \cdot, o) = \Dbase(\cdot, \cdot, o)$
% pointwise, so the sampler evaluates identical functions and returns identically
% distributed action chunks.
% \end{proof}

\begin{remark}
[Additive Property of Residual Denoisers]
\label{rem:residual-score}
Since $D \mapsto (D - x_\sigma)/\sigma^2$ is affine, writing $D_\phi = \Dbase + \Delta_\phi$ as in~\eqref{eq:adapted_denoiser} gives
% \vspace{-0.2cm}
\begin{equation}
    s_{D_\phi}(x_\sigma, \sigma, o)
    =
    s_{\Dbase}(x_\sigma, \sigma, o) + \sigma^{-2}\, \Delta_\phi(x_\sigma, \sigma, o),
    \label{eq:residual-score}
\end{equation}
so the probability-flow ODE~\eqref{eq:pf-ode} under $s_{D_\phi}$ is that of $\Dbase$ with the additional drift $-\Delta_\phi(x_\sigma, \sigma, o)/\sigma$. In particular, whenever $\Delta_\phi(\cdot, \cdot, o) \equiv 0$ for a given $o$, the sampler evaluates $D_\phi(\cdot,\cdot,o) = \Dbase(\cdot,\cdot,o)$, so $\pi_\phi(\cdot \mid o) = \pibase(\cdot \mid o)$.
\end{remark}

% NOTE: The two-stream architecture description moved, condensed, to Implementation Details with a pointer to Figure 1 (Lasse: figure/equations preferred over words).

We zero-initialize the output pathway of the adapter, following the convention for residual adapter modules~\cite{zhang2023adding}, so that $\Delta_\phi \equiv 0$ at initialization. By Remark~\ref{rem:residual-score}, training therefore starts from a policy that is exactly $\pibase$.

\para{Properties of Coordination via a Residual Denoiser.}
%The residual is more than an architectural convenience.
By Remark~\ref{rem:residual-score}, the adapter is an additive correction to the score field of the base policy at every noise level, and $D_\phi$ is plugged into the sampler of~\eqref{eq:pf-ode} with no change to the noise schedule or the solver.
Fig.~\ref{fig:residual-field} illustrates this view.
Because the correction acts on the vector field rather than on a sampled action, $\pi_\phi$ can modify the distribution of $\pibase$, unlike residual policy learning that offsets the output of a base controller~\cite{silver2018residual,johannink2019residual}.
As shown in Sec.~\ref{sec:experiments}, this residual adaptation of $\Dbase$ improves data efficiency since the frozen prior $\pibase$ already carries the basic skills that the collaborative task reuses.
Hence, the adapter needs to represent only the inter-agent couplings that the collaborative task requires. 
% As a result, the correction may require far less capacity than the base, which leaves it little room to overfit to the handful of collaborative demonstrations; we validate this data-efficiency property in \secref{sec:two_arm_results}.

% Lasse: let's just have this discussion in the related work section, not here.
% This denoiser-residual approach is akin to that of CoDi~\cite{peters2026coordinated}, which composes pretrained single-agent diffusion policies with a multi-agent cost, resulting in a similar residual term to the score field.
% In contrast to CoDi, however, we learn the residual correction from limited collaborative demonstrations rather than from a user-specified cost function.
%Moreover, note that the correction term \emph{reshapes an entire action distribution} rather than offsetting a single action, which distinguishes it from residual policy learning that adds a correction to the actions of a base controller~\cite{silver2018residual,johannink2019residual}: the adapted policy inherits the multi-modality of the prior and can sharpen, shift, or reweight its modes as the collaborative data demand.

\subsection{Mixed-Domain Training} \label{sec:mixed_training}

The residual adapter gives $\pi_\phi$ room to learn coordination.
However, when training this adapter, we must ensure that we still preserve the source-task behavior.
%As mentioned earlier, our goal is to learn multi-agent coordination while also preserving single-agent capabilities.
The residual formulation \eqref{eq:adapted_denoiser} allows us to implicitly capture this: the adapter $\Delta_\phi$ needs to learn to output zero in single-agent settings when no coordination is necessary while providing necessary non-zero corrections for collaborative tasks.
{In Fig.~\ref{fig:residual-field}, this means leaving the base field untouched on source-task observations (left) and correcting it only when a partner is in view (right).}
%This ensures that $\pi_\phi \approx \pibase$ and hence we utilize the base policy's capability in single-agent domains.
To achieve this, we train $\phi$ on both single-agent (source) and multi-agent (coordination) demonstrations simultaneously using a combination of two loss terms which we discuss next. %two data sources simultaneously, with a different loss on each.

\para{Policy-Distilled Replay.}
% A diffusion policy is a generative model of the behavior it was trained on, so rolling it out in its own environment produces samples of the very behavior we wish to preserve. 
Recall that, as per our problem formulation (\secref{sec:prob}), we do not have access to the source task training dataset.
Hence, before we can present our training objectives, we must first discuss how to generate suitable single-agent data for the adapter training.
To this end, we execute the frozen $\pibase$ in the source environment, following a receding-horizon loop, and record the resulting $M_S$ episodes of observations and executed actions, $\mathcal{D}_S = \{ \tau^S_j \}_{j=1}^{M_S}$ with $\tau^S_j = \{ (o_t, a_t) \}_{t=0}^{T-1}$. We call $\mathcal{D}_S$ \emph{policy-distilled replay}: it plays the role of rehearsal in continual learning~\cite{robins1995catastrophic,shin2017continual} with no expert effort.
% We specify the replay budget $M_S$ and the collection protocol we used in \secref{sec:experiment_settings}.
% TODO(review): the current implementation (scripts/distill_forwardonly_singlearm.py) keeps only rollouts that succeed at the source task; decide whether to commit to that success filter in the text here.

\para{Coordination Loss.}
On the collaborative demonstrations, we apply the denoising loss of~\eqref{eq:denoising-loss} to the adapted denoiser, $\mathcal{L}_{\mathrm{coord}}(\phi) = \mathcal{L}_{\mathrm{den}}(\phi; \mathcal{D}_C)$, with the same noise distribution $p(\sigma)$ and weighting $w(\sigma)$ as in the base policy's training and with gradients flowing only into $\phi$. Because all agents share the policy, we pool the {per-agent} action-observation-chunk tuples {$(o^i_t, a^i_{t:t+H})$} of all $N$ agents in $\mathcal{D}_C$ during training{, so no training sample contains another agent's observation}.

\para{Replay Loss.}
On the replay data, we penalize the residual itself rather than regressing $D_\phi$ onto the replayed actions, because by Remark~\ref{rem:residual-score}, a zero residual recovers $\pibase$ exactly. Therefore, it suffices to drive the residual to zero on source-task inputs, and we penalize it directly,
\begin{equation}
\mathcal{L}_{\mathrm{replay}}(\phi)
=
\mathbb{E}_{\mathcal{D}_S,\, \sigma,\, \epsilon}
\left[
    w(\sigma)
    \left\|
        \Delta_\phi(x + \sigma \epsilon, \sigma, o)
    \right\|_2^2
\right],
\label{eq:replay-loss}
\end{equation}
where the expectation runs over $(x, o) \sim \mathcal{D}_S$ with the same $\sigma$ and $\epsilon$ distributions as in~\eqref{eq:denoising-loss}.

\para{Combined Adaptation Objective.}
We train the adapter on the weighted sum of the two objectives, $\mathcal{L}(\phi) = \mathcal{L}_{\mathrm{coord}}(\phi) + \lambda_{\mathrm{ret}}\, \mathcal{L}_{\mathrm{replay}}(\phi)$, where the retention weight $\lambda_{\mathrm{ret}} > 0$ trades coordination gain against source-task retention. %: a larger value holds the residual closer to zero on source inputs at the price of a slower collaborative fit. We report the value we use, together with the sampling ratio between the two datasets, in \secref{sec:experiment_settings}.
This objective drives the residual to zero when the observation resembles the source task and makes it corrective when it shows a partner at work.
As a result, we do not require an explicit routing mechanism inside the adapter to switch between single-agent and multi-agent behavior.
% As we shall see in Sec.~\ref{sec:retention_results},  this formulation achieves source-task retention and collaborative success in \secref{sec:retention_results}.

\subsection{Deployment and Implementation Details} \label{sec:policy_architecture}

Since all agents share one set of weights, we instantiate a single copy of the frozen $\Dbase$ together with $\Delta_\phi$ and deploy it on every agent, each acting only on the RGB images from its own cameras. Every agent runs the pretrained policy's receding-horizon loop with $D_\phi$ as its denoiser and receives no signal about whether a partner is present (\secref{sec:prob}). Execution is fully decentralized: agents exchange no information and coordinate only through the physical workspace, as modeled in~\eqref{eq:dec-pomdp-dynamics}. Because the adapter is smaller than the backbone, it adds little inference cost over the pretrained policy.

Our base denoiser is a DiT-style transformer~\cite{peebles2023scalable} that consumes the noised chunk, the noise level, and visual tokens produced by an image encoder; \method{} touches this network only through its forward pass. We implement $\Delta_\phi$ as a small DiT-style decoder with the EDM preconditioning of~\cite{karras2022elucidating}. It is conditioned on two things. First, it uses the base encoder's token features through learned projections, in the spirit of side-tuning~\cite{zhang2020side}. Second, the raw observations are also processed through a lightweight convolutional network. This lets the adapter attend to partner cues that the frozen encoder, trained without partners in view, may discard. Figure~\ref{fig:fig1} (middle) depicts both networks. 
% In the next sections, we evaluate \method{} in simulation (\secref{sec:experiments}) and on hardware (\secref{sec:hardware_experiments}).

\section{Simulation Experiments} \label{sec:experiments}

    We conduct two-arm simulation experiments to answer three main questions.
    (Q1) Does \method achieve higher coordination success than training from scratch when multi-agent demonstrations are limited?
    (Q2) Does \method preserve source-task behavior?
    (Q3) How does the coordination-head size of \method affect coordination and source success when multi-agent demonstrations are limited?
    % (Q4) Do the performance characteristics of our method extend to a three-arm task?

    \subsection{Experiment Settings} \label{sec:experiment_settings}

        \para{Two-Arm Task.} 
        % As our main benchmark task, we study decentralized coordination of two robots simultaneously manipulating objects in a shared workspace.
        % We call this task \texttt{TwoArmPlaceWipe}.
        % In accordance with our problem formulation~(\secref{sec:prob}), each robot acts only from its local visual observations while sharing the workspace with its partner.
        We use \texttt{TwoArmPlaceWipe} as our main benchmark, where two robots coordinate from local visual observations while manipulating objects in a shared workspace.
        In the intended task sequence, Robot~1 lifts a handled tray, places it on one of two waiting areas, and returns it.
        Robot~2 picks up a sponge to wipe up the dirt under the tray and returns the sponge to the sponge pad.
        This task requires spatial and temporal coordination between the two arms in a shared space: Robot~1 must clear the dirt grid before Robot~2 can wipe it and must delay returning the tray until Robot~2 has completed the wipe. Any break in this sequence would likely lead to robot collision and task failure.
        The underlying frozen single-arm policy provides the place-and-return and wipe-and-return source behaviors.
        % We randomize the objects', pads', and dirt grid's initial locations across episodes to test generalization beyond the demonstrated configurations.
        We randomize the initial positions of objects, pads, and dirt across demonstrations and evaluation episodes to provide diverse task instances and evaluate generalization to configurations absent from the demonstrations.

        % \para{Three-Arm Task.} To stress-test our approach in a more complex setting, we also study a three-robot task, denoted \texttt{ThreeArmWipe}, in which two arms lift a tray by its two handles while a third arm wipes the dirt grid with a sponge.
        % The underlying frozen single-arm policy provides place, wipe, and tray-drag subtasks; these source subtasks are distinct from the robots' roles in the multi-agent task.
        % We randomize positions in the table plane. The tray, its return position, and the dirt grid share one offset; the sponge's initial position and each eligible return position vary independently.

        \para{Training Protocol.} We train using AdamW with a constant learning rate of $2\times10^{-4}$ and weight decay of $1\times10^{-4}$ for $900$k optimizer steps, saving exponential-moving-average-filtered checkpoints every $100$k steps.

        \para{Evaluation Protocol.} \label{sec:metrics} We evaluate each selected policy in both the multi-agent and single-arm domains. We report \emph{coordination success}, the fraction of multi-agent episodes satisfying the task-specific success criteria (described below), and \emph{source success}, the success rate on each original single-arm task. We select the checkpoint with the highest coordination success over $40$ evaluation episodes, then evaluate it on $200$ fresh multi-agent episodes. To assess source-task retention, we evaluate the same checkpoint on fresh episodes of the original single-arm tasks, keeping the coordination head active for our method, and compare source success with the unadapted base policy.
        We resample each model every $15$ environment steps. We use a $1{,}800$-step episode cap as a hard cutoff for these samples. %for the two-arm evaluations and a $3{,}000$-step cap for the three-arm coordination evaluations. 
         Table~\ref{tab:twoarm-primary} reports the completed multi-agent evaluations. %and~\ref{tab:threearm-primary} report the completed multi-agent evaluations.

        \para{Coordination Success Criteria.} This task uses coordination success as the main evaluation metric.
        We count an episode as successful when the following conditions hold: the tray is less than $1.5$\,cm from the return position along both the $x$ and $y$ axes; the sponge is offset no more than $8$\,cm along $x$ and $6$\,cm along $y$ from its target; and wipe coverage reaches at least $60\%$. % ($24$ of $40$ dirt tiles).

        % For the \texttt{ThreeArmWipe} task, we count an episode as successful when the following conditions hold: the tray is tilted no more than~$20^\circ$ and comes within $10$~cm of the waiting area before being placed within $8$\,cm of its return position; the sponge reaches at least $80\%$ wipe coverage before being placed within $10$\,cm of its return location.

        \noindent\textbf{Training Data and Budgets.} Since collecting coordinated demonstrations requires operating multiple robots simultaneously, it is desirable to achieve policy adaptation with as few multi-agent demos as possible.
        Therefore, we study policy performance under limited multi-agent data budgets.
        % We evaluate three adaptation-data budgets for each simulation task: $20$, $40$, and $60$ coordinated multi-agent demonstrations.
        For the \texttt{TwoArmPlaceWipe} task, mixed-data training includes $20$, $40$, and $60$ single-arm demonstrations, respectively. 
        % for the \texttt{ThreeArmWipe} task, mixed-data training includes $40$, $80$, and $120$ single-arm demonstrations, respectively.
        These single-arm demonstrations are policy-distilled rollouts from the pretrained policy and are entirely separate from the 600 expert single-arm demonstrations used to train the base policy.
        Multi-agent-only training reuses the same multi-agent demonstrations.
        %The tables report the two counts separately; neither includes the data originally used to pretrain the base policy.
        % For each task and budget $K\in\{5,10,15\}$, we collected $K$ multi-agent demonstrations per configuration, together with policy-distilled single-arm rollouts of the source tasks.
        % For \texttt{TwoArmPlaceWipe}, configurations correspond to combinations of tray destination and sponge return pad, and we collected $K$ single-arm rollouts per single-arm configuration. This yielded $20+20$, $40+40$, and $60+60$ multi-arm and single-arm files, respectively. For \texttt{ThreeArmWipe}, larger datasets retained the lower-budget demonstrations and source rollouts.

        % When training with both multi-agent demonstrations and single-arm source rollouts, we use a fixed $1{:}1$ balance between single-agent and multi-agent demonstrations: each optimizer step receives $256$ multi-arm samples and $256$ single-arm samples, regardless of the relative dataset sizes.
        % When training only with multi-agent demonstrations, each step receives $256$ multi-arm samples and no single-arm samples.
        % Within each domain, hierarchical sampling balances task/mode, rollout, agent, and then timestep.
        For mixed-data training, each optimizer update processes a batch of $512$ samples, split equally between multi-agent and single-arm data. Multi-agent-only training uses $256$ multi-agent samples per update.

        \begin{figure}[t]
            \centering
            \includegraphics[width=1.0\linewidth]{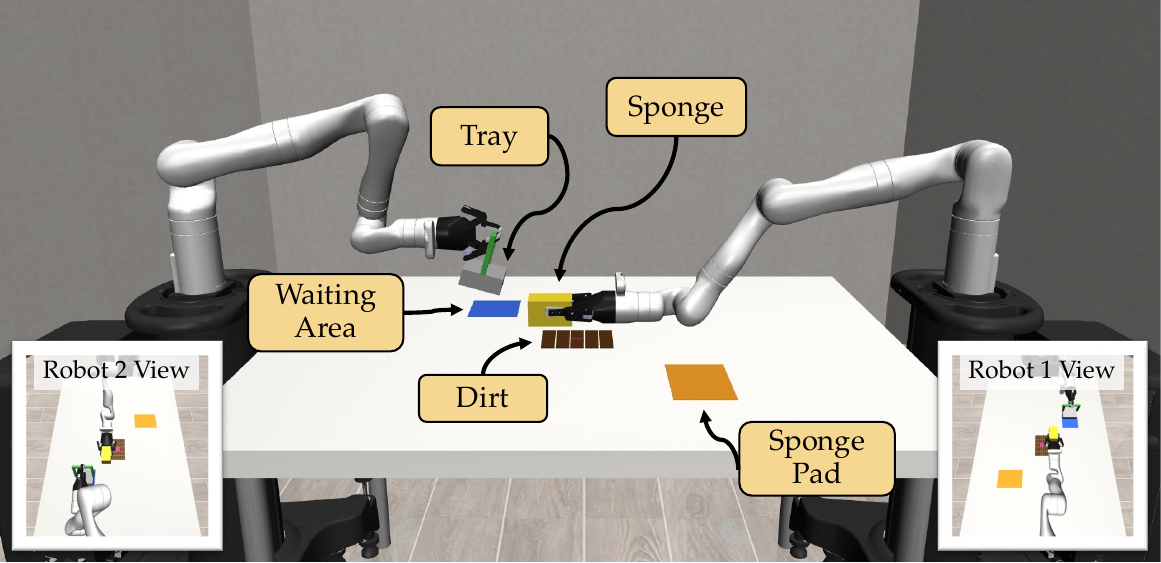}
            \caption{In the \texttt{TwoArmPlaceWipe} simulated task, one robot places and returns a tray while another wipes up dirt.}
            \label{fig:simulation-tasks}
            \vspace{-0.5cm}
        \end{figure}

    \subsection{Baselines} \label{sec:baselines}

        Our evaluation compares \method with three baselines to examine how training from scratch or fine-tuning the pretrained policy affects coordination success and source-task retention under limited multi-agent data.
        We condition all methods only on RGB image observations from their own shoulder camera. Hence, all approaches are \emph{decentralized} at test time.

        \para{From Scratch (\textsc{FS}).} To test the benefit of reusing pretrained single-arm behavior, we train a baseline policy from scratch using the same multi-agent demonstrations and policy-distilled single-arm rollouts that we use to train the coordination head.
        Its parameter count closely matches the sum of the coordination-head and frozen single-arm-policy parameter counts; we detail the matching procedure in Section~\ref{sec:ablations}.

        \para{Fine-tuning with Mixed Data (\textsc{FT-mixed}).} To test whether updating the pretrained policy directly is preferable to \method's strategy (freezing the pretrained policy and adding a coordination head), we set up a baseline that fine-tunes the pretrained policy using both multi-agent and single-agent data.
        % This baseline optimizes all $17{,}446{,}215$ parameters of the pretrained single-arm policy combined with exponential-moving-average (EMA) filtering of the policy weights.
        Both fine-tuning baselines optimize all $17{,}446{,}215$ parameters of the pretrained single-arm policy without adding modules or partner-specific inputs.
        %Both finetuning baselines follow the training and evaluation protocol in Section~\ref{sec:experiment_settings}.
        %At each optimizer step, we will concatenate $256$ samples from each domain into a $512$-sample batch and optimize a unified mean loss.

        \para{Fine-tuning with Multi-Agent Data Only (\textsc{FT-multi}).} To test whether source replay is needed during adaptation, we also include a variant of FT-mixed that fine-tunes the pretrained policy using \emph{only} the multi-agent demonstrations. %We use the same multi-agent demonstrations and the same number of demonstrations per mode as the other methods.

    \subsection{Two-Arm Low-Data Coordination} \label{sec:two_arm_results}

        To test our method's multi-agent data efficiency, we compare all methods under limited multi-agent data budgets.

        \para{Results.}
        Tab.~\ref{tab:twoarm-primary} reports coordination success across three data budgets.
        % \method exceeds \textsc{FS} by $21.5$, $43.5$, and $28.0$ percentage points at five, ten, and fifteen demonstrations per mode, respectively.
        \method exceeds \textsc{FS} by $21.5$, $43.5$, and $28.0$ percentage points at $20$, $40$, and $60$ multi-agent demonstrations, respectively.
        \textsc{FT-mixed} achieves $33.0\%$, $59.5\%$, and $63.5\%$ at these budgets, exceeding \textsc{FS} but remaining below \method.
        \textsc{FT-multi} achieves $29.5\%$, $50.5\%$, and $58.5\%$, below \textsc{FT-mixed} at every budget.

        \para{Key Takeaway.}
        These results answer Q1: adapting a frozen single-arm policy with a coordination head outperforms learning from scratch when multi-agent data is limited.

        % \begin{table}[t]
        %     \centering
        %     \caption{\texttt{TwoArmPlaceWipe} coordination success (\%, $\uparrow$) across demonstration budgets.}
        %     \label{tab:twoarm-primary}
        %     \setlength{\tabcolsep}{3pt}
        %     \begin{tabular}{cclcc}
        %         \toprule
        %         \shortstack{Multi-agent\\demos} &
        %         \shortstack{Single-arm\\demos} &
        %         Method & Step & \shortstack{Coordination\\success $\uparrow$} \\
        %         \midrule
        %         20 & 20 & \method    & $500$k & 35.0 \\
        %         20 & 20 & \textsc{FS}       & $200$k & 13.5 \\
        %         20 & 20 & \textsc{FT-mixed} & $200$k & 33.0 \\
        %         20 & 0  & \textsc{FT-multi} & $200$k & 29.5 \\
        %         40 & 40 & \method    & $300$k & 71.5 \\
        %         40 & 40 & \textsc{FS}       & $900$k & 28.0 \\
        %         40 & 40 & \textsc{FT-mixed} & $400$k & 59.5 \\
        %         40 & 0  & \textsc{FT-multi} & $300$k & 50.5 \\
        %         60 & 60 & \method    & $500$k & 87.0 \\
        %         60 & 60 & \textsc{FS}       & $400$k & 59.0 \\
        %         60 & 60 & \textsc{FT-mixed} & $500$k & 63.5 \\
        %         60 & 0  & \textsc{FT-multi} & $200$k & 58.5 \\
        %         \bottomrule
        %     \end{tabular}
        % \end{table}
        \begin{table}[t]
            \centering
            \caption{\texttt{TwoArmPlaceWipe} coordination success (\%, $\uparrow$) across demonstration budgets.
            Bold denotes the best result within each data regime.}
            \label{tab:twoarm-primary}
            \setlength{\tabcolsep}{3pt}
            \begin{tabular}{cclcc}
                \toprule
                \shortstack{Multi-agent\\demos} &
                \shortstack{Single-arm\\demos} &
                Method &
                Checkpoint &
                \shortstack{Coordination\\success $\uparrow$} \\
                \midrule
        
                20 & 20 & \method
                & $500$k & \textbf{35.0} \\
                20 & 20 & \textsc{FS}
                & $200$k & 13.5 \\
                20 & 20 & \textsc{FT-mixed}
                & $200$k & 33.0 \\
                20 & 0 & \textsc{FT-multi}
                & $200$k & 29.5 \\
        
                \midrule
        
                40 & 40 & \method
                & $300$k & \textbf{71.5} \\
                40 & 40 & \textsc{FS}
                & $900$k & 28.0 \\
                40 & 40 & \textsc{FT-mixed}
                & $400$k & 59.5 \\
                40 & 0 & \textsc{FT-multi}
                & $300$k & 50.5 \\
        
                \midrule
        
                60 & 60 & \method
                & $500$k & \textbf{87.0} \\
                60 & 60 & \textsc{FS}
                & $400$k & 59.0 \\
                60 & 60 & \textsc{FT-mixed}
                & $500$k & 63.5 \\
                60 & 0 & \textsc{FT-multi}
                & $200$k & 58.5 \\
        
                \bottomrule
            \end{tabular}
            \vspace{-0.3cm}
        \end{table}
        %Counts refer to adaptation data only; single-arm demonstrations are policy-distilled rollouts and exclude base-policy pretraining data.
    \subsection{Source-Task Retention} \label{sec:retention_results}

        Coordination gains may come at the cost of single-agent skills. We therefore evaluate the same checkpoints as in Table~\ref{tab:twoarm-primary} on the original place-and-return and wipe-and-return tasks of the source domain.
        % \begin{table}[t]
        %     \centering
        %     \caption{Source-task success (\%, $\uparrow$) after two-arm adaptation.}
        %     \label{tab:source-retention}
        %     \small
        %     \setlength{\tabcolsep}{2pt}
        %     \begin{tabular}{@{}cclccc@{}}
        %         \toprule
        %         & & & \multicolumn{3}{c}{Source success $\uparrow$} \\
        %         \cmidrule(lr){4-6}
        %         \shortstack{Multi-agent\\demos} &
        %         \shortstack{Single-arm\\demos} &
        %         Method & \shortstack{Place-\\return} & Wipe &
        %         Combined \\
        %         \midrule
        %         --- & --- & Base policy & 100.0 & 100.0 & 100.0 \\
        %         \midrule
        %         20 & 20 & \method    & 99.0  & 95.0 & 97.0 \\
        %         20 & 20 & \textsc{FS}       & 56.0  & 9.0  & 32.5 \\
        %         20 & 20 & \textsc{FT-mixed} & 55.0  & 37.0 & 46.0 \\
        %         20 & 0  & \textsc{FT-multi} & 0.0 & 40.0 & 20.0 \\
        %         40 & 40 & \method    & 100.0 & 95.0 & 97.5 \\
        %         40 & 40 & \textsc{FS}       & 75.0  & 14.0 & 44.5 \\
        %         40 & 40 & \textsc{FT-mixed} & 84.0  & 81.0 & 82.5 \\
        %         40 & 0  & \textsc{FT-multi} & 0.0 & 68.0 & 34.0 \\
        %         60 & 60 & \method    & 99.0  & 95.0 & 97.0 \\
        %         60 & 60 & \textsc{FS}       & 76.0  & 46.0 & 61.0 \\
        %         60 & 60 & \textsc{FT-mixed} & 83.0  & 82.0 & 82.5 \\
        %         60 & 0  & \textsc{FT-multi} & 0.0 & 50.0 & 25.0 \\
        %         \bottomrule
        %     \end{tabular}
        % \end{table}
        \begin{table}[t]
            \centering
            \caption{Source-task success (\%, $\uparrow$) after two-arm adaptation.
            %Bold denotes the best result within each data regime.
            }
            \label{tab:source-retention}
            \small
            \setlength{\tabcolsep}{2pt}
            \renewcommand{\arraystretch}{0.98}
            \begin{tabular}{@{}ccclccc@{}}
                \toprule
                & \multicolumn{2}{c}{Single-arm demos} & 
                & \multicolumn{3}{c}{Source success $\uparrow$} \\
                \cmidrule(lr){2-3}
                \cmidrule(lr){5-7}
                \shortstack{Multi-arm\\demos} &
                Expert &
                \shortstack{Distilled\\rollouts} &
                Method &
                \shortstack{Place-\\return} &
                Wipe &
                Combined \\
                \midrule
        
                0 & 400 & 0 & Base policy
                & 100.0 & 94.0 & 97.0 \\
        
                \midrule
        
                20 & 0 & 20 & \method
                & \textbf{99.0} & \textbf{95.0} & \textbf{97.0} \\
                20 & 0 & 20 & \textsc{FS}
                & 56.0 & 9.0 & 32.5 \\
                20 & 0 & 20 & \textsc{FT-mixed}
                & 55.0 & 37.0 & 46.0 \\
                20 & 0 & 0 & \textsc{FT-multi}
                & 0.0 & 40.0 & 20.0 \\
        
                \midrule
        
                40 & 0 & 40 & \method
                & \textbf{100.0} & \textbf{95.0} & \textbf{97.5} \\
                40 & 0 & 40 & \textsc{FS}
                & 75.0 & 14.0 & 44.5 \\
                40 & 0 & 40 & \textsc{FT-mixed}
                & 84.0 & 81.0 & 82.5 \\
                40 & 0 & 0 & \textsc{FT-multi}
                & 0.0 & 68.0 & 34.0 \\
        
                \midrule
        
                60 & 0 & 60 & \method
                & \textbf{99.0} & \textbf{95.0} & \textbf{97.0} \\
                60 & 0 & 60 & \textsc{FS}
                & 76.0 & 46.0 & 61.0 \\
                60 & 0 & 60 & \textsc{FT-mixed}
                & 83.0 & 82.0 & 82.5 \\
                60 & 0 & 0 & \textsc{FT-multi}
                & 0.0 & 50.0 & 25.0 \\
        
                \bottomrule
            \end{tabular}
        % \vspace{-0.3cm}
        \end{table}
        % Table~\ref{tab:source-retention} shows that \method achieves macro-average source success of $97.0\%$--$97.5\%$, compared with $32.5\%$--$61.0\%$ for \textsc{FS}. The largest task-specific gap occurs on wipe-and-return. The unadapted base policy achieves $100.0\%$ on both tasks, with $20$ evaluation episodes per task compared with $100$ for the adapted policies.
        
        \para{Results.} Table~\ref{tab:source-retention} shows that \method achieves combined source success of $97.0\%$--$97.5\%$, compared with $32.5\%$--$61.0\%$ for \textsc{FS}.
        \textsc{FT-mixed} achieves $46.0\%$, $82.5\%$, and $82.5\%$ at $20$, $40$, and $60$ multi-agent demonstrations, respectively, also remaining below \method.
        \textsc{FT-multi} achieves $20.0\%$, $34.0\%$, and $25.0\%$, below all three other methods at each budget.
        The largest task-specific gap between \method and \textsc{FS} occurs on wipe-and-return.
        The unadapted base policy achieves $100.0\%$ on place-return and $94\%$ on the wipe task, with $100$ evaluation episodes per source-task for the adapted policies and the base policy.

        \para{Key Takeaway.}
        These results answer Q2: despite the addition of the coordination head, the adapted policy maintains high success on both original single-arm tasks.
        % TODO: Add the finetuning comparison. Use matched episode-level outcomes
        % if quantifying adaptation effects on the shared evaluation seeds.

    \subsection{Ablation Studies} \label{sec:ablations}

        Coordination-head size may affect both learning from limited demonstrations and source-task retention. We examine these effects by varying the head size while keeping the source policy frozen (Q3).

        % We evaluate heads of four different sizes on \texttt{TwoArmPlaceWipe} using five demonstrations per mode.
        % We hold the base policy, data, observation interface, and training and evaluation protocol fixed across sizes.
        % For each size, we match an \textsc{FS} policy to the combined base-plus-head parameter count by varying only its transformer and feed-forward widths.
        % All other \textsc{FS} architecture choices remain fixed, and parameter counts differ from their targets by less than $0.0013\%$. We trained every \textsc{FS} parameter from scratch.

        % Table~\ref{tab:head-size} reports the policy performance in terms of coordination success and average source success over the place-and-return and wipe-and-return source tasks across the four model sizes. The main two-arm and three-arm experiments use the largest head.
        % \todo{add one sentence of interpretation of high-level trend(s) linking back to Q3 as a final sentence as we have for other results}
        \para{Setup.}
        We evaluate heads of four different sizes on \texttt{TwoArmPlaceWipe} using $20$ multi-agent and $20$ single-arm demonstrations.
        We hold the base policy, data, observation interface, and training and evaluation protocol fixed across all sizes.
        For each size, we match an \textsc{FS} policy to the combined base-plus-head parameter count by varying only its transformer and feed-forward widths.
        Through this matching strategy, parameter counts are within $0.0013\%$ of the desired base-plus-head parameter count.
        All other \textsc{FS} architecture choices remain fixed.
        %We trained every \textsc{FS} parameter from scratch.

        \para{Results.}
        % Table~\ref{tab:head-size} lists the parameter counts. We measure coordination success and combined source success, weighting place-and-return and wipe-and-return equally.
        % We find that coordination success for \method increases from $25.0\%$ with the XS head to $35.0\%$ with the L head, while combined source success remains above $96.5\%$. Each variant achieves higher coordination success and source success than its parameter-matched \textsc{FS} policy.
        % Therefore, the main two-arm and three-arm experiments use the largest head.
        Coordination success for \method increases from $25.0\%$ with the XS head to $35.0\%$ with the L head, while every reported head retains $97.0\%$ combined source success. Each coordination-head variant achieves higher coordination and source success than its parameter-matched \textsc{FS} policy. We therefore use the L head in the main two-arm experiments. %and three-arm experiments.

        % \begin{table}[t]
        %     \centering
        %     \caption{Effect of coordination-head size on coordination and source success (\%, $\uparrow$) in \texttt{TwoArmPlaceWipe}.}
        %     \label{tab:head-size}
        %     \setlength{\tabcolsep}{3pt}
        %     \begin{tabular}{clcccc}
        %         \toprule
        %         Size & Method & Trainable (M) & Total (M) &
        %         \shortstack{Coordination\\success $\uparrow$} &
        %         \shortstack{Combined\\source success $\uparrow$} \\
        %         \midrule
        %         XS & \method     & 0.435  & 17.881 & 25.0 & 97.0 \\
        %         XS & \textsc{FS} & 17.881 & 17.881 & 12.0 & 34.5 \\
        %         S  & \method     & 1.630  & 19.077 & 27.0 & 97.0 \\
        %         S  & \textsc{FS} & 19.076 & 19.076 & 10.0 & 37.5 \\
        %         M  & \method     & 3.369  & 20.815 & 33.5 & 97.0 \\
        %         M  & \textsc{FS} & 20.815 & 20.815 & 10.5 & 39.5 \\
        %         L  & \method     & 7.400  & 24.846 & 35.0 & 97.0 \\
        %         L  & \textsc{FS} & 24.846 & 24.846 & 13.5 & 32.5 \\
        %         \bottomrule
        %     \end{tabular}
        % \end{table}
        \begin{table}[t]
            \centering
            \caption{Effect of coordination-head size on coordination and source-task
            success (\%, $\uparrow$) in \texttt{TwoArmPlaceWipe}.
            %Bold denotes the best result in each performance metric; ties are bolded.
            }
            \label{tab:head-size}
            \setlength{\tabcolsep}{3pt}
            \begin{tabular}{clcccc}
                \toprule
                Size &
                Method &
                \shortstack{Total\\params. (M)} &
                \shortstack{Trainable\\fraction (\%)} &
                \shortstack{Coordination\\success $\uparrow$} &
                \shortstack{Source\\success $\uparrow$} \\
                \midrule
        
                XS & \method
                & 17.881 & 2.4 & 25.0 & \textbf{97.0} \\
                XS & \textsc{FS}
                & 17.881 & 100.0 & 12.0 & 34.5 \\
        
                \midrule
        
                S & \method
                & 19.077 & 8.5 & 27.0 & \textbf{97.0} \\
                S & \textsc{FS}
                & 19.076 & 100.0 & 10.0 & 37.5 \\
        
                \midrule
        
                M & \method
                & 20.815 & 16.2 & 33.5 & \textbf{97.0} \\
                M & \textsc{FS}
                & 20.815 & 100.0 & 10.5 & 39.5 \\
        
                \midrule
        
                L & \method
                & 24.846 & 29.8 & \textbf{35.0} & \textbf{97.0} \\
                L & \textsc{FS}
                & 24.846 & 100.0 & 13.5 & 32.5 \\
        
                \bottomrule
            \end{tabular}
            \vspace{-0.5cm}
        \end{table}
        
        \para{Key Takeaway.}
        These results address Q3: larger coordination heads achieve higher coordination success in this comparison, while all four sizes maintain high source success.

\section{Hardware Experiments} \label{sec:hardware_experiments}

% Finally, to examine whether the performance advantage of our approach observed in simulation transfers to the real world, we evaluate our approach against the strongest baseline variants of Sec.~\ref{sec:experiments} in a quantitative hardware experiment.
Finally, we evaluate whether the simulation results transfer to hardware using the strongest baseline variants from Sec.~\ref{sec:experiments}.

\subsection{Setup}

\begin{figure}[t]
        \centering
        \includegraphics[width=1.0\linewidth]{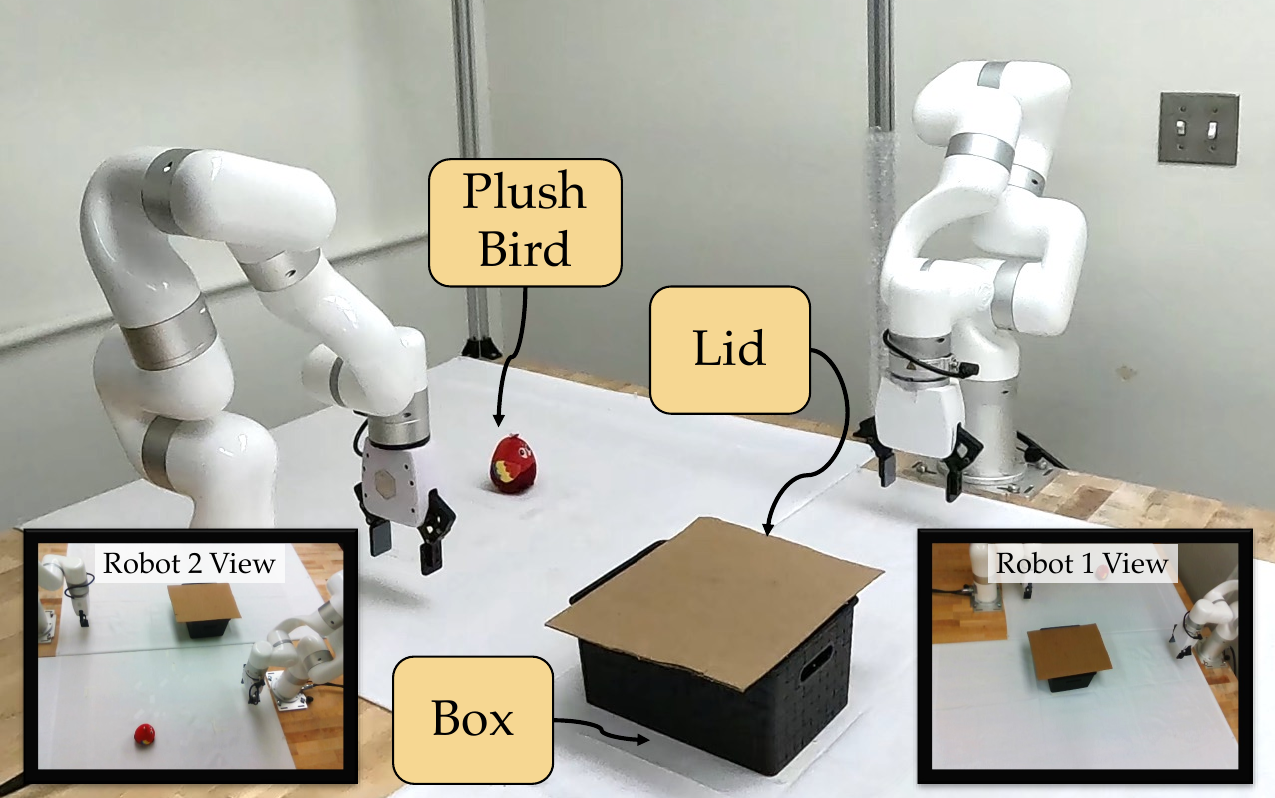}
        \caption{Hardware setup with two xArm7 robots, a lidded box, and a plush bird. One robot removes the lid, the other places the bird in the box, and the first robot replaces the lid.}
        \label{fig:hardware-setup}
        \vspace{-0.4cm}
    \end{figure}

\para{Task Description.}
In our hardware experiments, two 7-dof xArm7 robots are tasked with placing a plush bird into a lidded box.
To achieve this, Robot~1 must temporarily remove the lid before Robot~2 places the bird in the open box.
We consider the task successfully completed if the bird is inside the box and the lid covers more than $50\%$ of the box while remaining in place without robot support.
Fig.~\ref{fig:hardware-setup} shows the setup for this experiment.

\para{Baselines.}
As baselines, we consider the strongest baselines from our simulation experiments~(\secref{sec:experiments}): \textsc{FT-mixed} as the strongest fine-tuning variant, and \textsc{FS} as a baseline that trains on all available data from scratch rather than fine-tuning.

\para{Demonstration Data.}
We train each method using $69$ two-arm demonstrations and $68$ policy-distilled single-arm demonstrations.
We pretrain the single-arm policy used by \method and \textsc{FT-mixed} on a total of $332$ single-arm demonstrations of three source tasks: lid removal, lid replacement, and bird-pick-place.
%We reset objects to approximately the same initial positions across trials, with slight variation in bird placement.
%We intentionally limit initial-state variation to focus the evaluation on physical coordination.
Trials have a $900$-timestep cutoff.

    % FT-multi was not evaluated on hardware.

\para{Implementation Details.}
Each robot receives observations from its own shoulder-view RGB camera.
Both the planning and execution horizons are $20$ timesteps.
We train all models for $25$k gradient steps.
All other hyperparameters match the nominal setting of Sec.~\ref{sec:experiment_settings}.
    
    \subsection{Results}\label{sec:hardware_two_arm_results}
    % Completing the box task requires the robots to coordinate an ordered sequence: the lid must be removed before the bird can be placed inside.
    % At the same time, the multi-agent data budget is small relative to the size of the joint action space of two 7-dof robots.
    %The limited demonstration budget tests each method's ability to learn to coordinate an ordered sequence while retaining source skills.
    % Hence, this task tests each method's ability to learn coordination from limited multi-agent demonstrations while retaining the source-task capabilities.
    Tab.~\ref{tab:hardware-coordination} reports the results of our hardware experiments.

    \para{Coordination.}
    %We evaluate \method, \textsc{FS}, and \textsc{FT-mixed} on $20$ coordination trials each. 
    Across $20$ multi-agent trials per method, we find that \method and \textsc{FT-mixed} each succeed in $18/20$ trials ($90\%$), compared with $13/20$ ($65\%$) for \textsc{FS}.
    Our method matches the observed coordination success of full-policy fine-tuning with mixed data (\textsc{FT-mixed}) and exceeds training from scratch (\textsc{FS}) by $25$ percentage points. % on this task.

    % \begin{table}[t]
    %     \centering
    %     \caption{Hardware coordination and source success. Each method uses $69$ two-arm and $68$ policy-distilled single-arm adaptation demonstrations, excluding base-policy pretraining data. Coordination success uses $20$ trials; each source task uses $10$. Entries report successful trials and their denominators ($\uparrow$). Bold entries indicate the highest observed success in each column.}
    %     \label{tab:hardware-coordination}
    %     \label{tab:hardware-retention}
    %     \setlength{\tabcolsep}{3pt}
    %     \begin{tabular}{lcccc}
    %         \toprule
    %         & & \multicolumn{3}{c}{Source success $\uparrow$} \\
    %         \cmidrule(lr){3-5}
    %         Method & \shortstack{Coordination\\success $\uparrow$} &
    %         \shortstack{Lid\\removal} & \shortstack{Lid\\replacement} &
    %         \shortstack{Bird\\pick-place} \\
    %         \midrule
    %         \method & \textbf{18/20} & \textbf{10/10} & \textbf{10/10} & \textbf{10/10} \\
    %         \textsc{FS} & 13/20 & 2/10 & \textbf{10/10} & \textbf{10/10} \\
    %         \textsc{FT-mixed} & \textbf{18/20} & 7/10 & 9/10 & 9/10 \\
    %         \bottomrule
    %     \end{tabular}
    % \end{table}
    \begin{table}[t]
        \centering
        \caption{Hardware coordination and source success. % Each method uses $69$ two-arm and $68$ policy-distilled single-arm adaptation demonstrations, excluding base-policy pretraining data. Coordination success uses $20$ trials; each source task uses $10$. Entries report successful trials and their denominators ($\uparrow$). Bold entries indicate the highest observed success in each column.
        }
        \label{tab:hardware-coordination}
        \label{tab:hardware-retention}
        % \setlength{\tabcolsep}{3pt}
        % \begin{tabular}{lcccc}
        \setlength{\tabcolsep}{3pt}
        \renewcommand{\arraystretch}{0.99}
        \begin{tabular}{lcccc}
            \toprule
            & & \multicolumn{3}{c}{Source success $\uparrow$} \\
            \cmidrule(lr){3-5}
            Method & \shortstack{Coordination\\success $\uparrow$} &
            \shortstack{Lid\\removal} & \shortstack{Lid\\replacement} &
            \shortstack{Bird\\pick-place} \\
            \midrule
            Base policy & -- & 40/40 & 40/40 & 40/40 \\
            \midrule
            \method & \textbf{18/20} & \textbf{19/20} & \textbf{20/20} & \textbf{20/20} \\
            \textsc{FS} & 13/20 & 9/20 & {19/20} & {12/20} \\
            \textsc{FT-mixed} & \textbf{18/20} & 11/20 & 19/20 & 17/20 \\
            \bottomrule
        \end{tabular}
        \vspace{-0.4cm}
    \end{table}

   % \subsection{Source Skill Retention}\label{sec:hardware_retention}
    %As in simulation, we use the same policies for coordination and source-task evaluation, keeping the coordination head active for \method. 

    \para{Source Skill Retention.}
    % Finally, using $20$ trials of each source task (lid removal, lid replacement, and bird-pick-place), we evaluate source skill retention of the \emph{adapted} policy learned by each method to determine whether physical coordination gains come at the cost of those skills. 
    We evaluate each adapted policy on $20$ trials of each source task.
    \method succeeds in all $20$ trials of lid replacement and bird-pick-place, and succeeds in $19$ out of $20$ trials of lid removal.
    \textsc{FS} succeeds in $19/20$ lid replacement trials, $12/20$ bird-pick-place trials, and $9/20$ lid removal trials. \textsc{FT-mixed} succeeds in $11/20$, $19/20$, and $17/20$ trials, respectively.
    In summary, our method ties for the highest observed coordination success while achieving the highest observed success on all three source tasks. The source-task evaluation distinguishes it from \textsc{FT-mixed}, despite their equal coordination success.

\section{Conclusion} \label{sec:conclusion}

We study adaptation of pretrained single-arm base policies to coordinate on demand, i.e., to achieve decentralized multi-agent coordination while retaining the ability to also act independently.
We study this problem under two key constraints: limited access to multi-agent coordination demonstrations and no access to the expert demonstrations underpinning the pretrained single-arm policy.
Our method, \method, freezes the pretrained policy and trains a residual coordination head using both multi-agent demonstrations and policy-distilled single-arm replay data. In simulation, \method improves coordination success over parameter-matched training from scratch by up to 43.5 percentage points and over full-policy fine-tuning with mixed data by up to 23.5 percentage points. It also improves combined source-task success by up to 64.5 and 51.0 percentage points over those baselines, respectively.
Hardware experiments on a collaborative manipulation task between two 7-dof xArm7 robots support these findings.
Future work will extend the approach to more diverse tasks and robot platforms.

% In extensive simulation experiments with varying demonstration budgets, \method achieves higher coordination success than from-scratch training or fine-tuning baselines while also retaining single-agent skills.
% This gap is largest at small data budgets, highlighting the utility of our approach for sample-efficient coordination on demand.
% Hardware experiments of a collaborative manipulation task between two 7-dof xArm7 robots validate this finding.
% Future work will extend the approach to more diverse tasks and robot platforms.

\FloatBarrier
\bibliographystyle{IEEEtran}
\bibliography{references}

\end{document}